\documentclass{article}
\usepackage{ijcai21}
\usepackage{times}

\usepackage[utf8]{inputenc}
\usepackage[inline]{enumitem}
\usepackage[hidelinks]{hyperref}
\usepackage{amsmath}
\usepackage{listings}

\title{An Automated Framework for Supporting Data-Governance Rule Compliance in Decentralized MIMO Contexts}
\author{
    Rui Zhao
    \affiliations
    University of Edinburgh
    \emails
    rui.zhao@ed.ac.uk
}

\begin{document}

\maketitle

\begin{abstract}
 We propose Dr.Aid, a logic-based AI framework for automated compliance checking of data governance rules over data-flow graphs. The rules are modelled using a formal language based on situation calculus and are suitable for decentralized contexts with multi-input-multi-output (MIMO) processes. Dr.Aid models \emph{data rules} and \emph{flow rules} and checks compliance by reasoning about the propagation, combination, modification and application of data rules over the data flow graphs. Our approach is driven and evaluated by real-world datasets using provenance graphs from data-intensive research. 
\end{abstract}

\section{Introduction}

The significant accomplishments of Artificial Intelligence and Data Science in the recent years have highlighted the value and importance of combining data sets from multiple sources, particularly in areas of global interest, such as healthcare. As we deal with such large collections of personal, sensitive, or otherwise valuable data, issues of data privacy, ownership and governance gain dominance.

Data sharing is typically bound by data-governance rules or data-use policies (\emph{policies} for short) established by the data provider and adopted by data users. However, compliance checking is usually performed manually, in a time consuming and error prone way. This diminishes the trust of the data provider in the users' ability to comply with the agreed rules. Coupled with increasingly strict legislation for responsible data sharing, data providers are reluctant to release any data with the slightest risk of public objections. This has created a polarization in data-governance rules: data is either released freely as \emph{open data} or under strict barriers to access, such as restricted data safe havens with controlled access, strict governance rules, meticulous and tedious application processes, supervision, etc. This has a direct impact on establishing and scaling collaborations for AI and data-intensive research across organisations and jurisdictions.

We propose the automation of compliance checking of policies using a symbolic AI agent. The agent facilitates enforcement of the desired policies, providing peace of mind to both data providers and users. It can also provide broader support, for instance by automatically deriving policies associated with data products and providing reminders for required actions (such as reporting data use at specific milestones).

More specifically, we envision a logic-based formalisation of data-governance rules coupled with an automated reasoning mechanism that can track and derive policies through every stage of the data-use workflow. This includes propagating data-governance rules from one or more data inputs to the corresponding outputs accordingly, merging policies from different upstreams, modifying the policies to reflect changes in the underlying data, and checking the application of policies in the current processes. 

With this goal, we have developed a formal language based on situation calculus to model data-governance rules and their propagation and a corresponding framework, \emph{Dr.Aid} (Data Rule Aid), to perform compliance reasoning on data-flow graphs via logic-based querying. We are evaluating our approach using real-world data-provenance graphs (a standardised representation of actions on data) from cyclone tracking and computational seismology applications. 

%
Prior research has been developed under restricted assumptions, such as having a single context, a single stage of processing, propagating rules unchanged without reflecting modifications in the data
~\cite{elnikety_thoth:_2016}, or assuming linear processing~\cite{pasquier_camflow:_2017}. 
Dr.Aid addresses all of these issues as they arise in practice.
%
%
%
It is explicitly designed for MIMO processes and multi-staged data flows, and dynamically updates rules to reflect data modifications.

\section{Approach}

Our framework, Dr.Aid, checks compliance with data-governance rules (policies)  suitable for decentralized MIMO data processing, using a formal language and semantics. It enables the modelling of data-use policies for decentralized MIMO contexts, and performs logical reasoning to deliver information about policies for specific workflow runs.

Our formal rule language consists of two inter-operating parts: \emph{data rules} and \emph{flow rules}~\cite{zhao_towards_2019}. 
\emph{Data rules} encode data-governance rules for multi-staged processing, e.g.~``\emph{users must properly acknowledge the data providers}''; \emph{flow rules} represent the changes of data rules in each process as a result of data transportation and transformation, e.g.~``\emph{column 3 from input 1 is changed to column 2 on output 2}''.
Thus, the flow rules specify how data rules are propagated and transformed from inputs to outputs for each process; data rules are propagated between processes by following the data-flow graph. The reasoner combines these two forms of rule handling for each workflow enactment.

The two main elements of our data rules are \emph{attributes}, each of which is a triple $(N, T, V)$ of a name $N$, a type $T$ and a value $V$ describing properties of the data, and \emph{obligations}, each of which is a triple $(OD, VB, AC)$ consisting of an obligation definition $OD$ (the obligated action to perform upon activation), a validity binding $VB$ (describing additional applicability constraints), and an activation condition $AC$ (the triggering condition). The attribute type $T$ and the obligated action type of $OD$ can refer to external definitions through a semantic approach, thus allowing the rule to unambiguously interoperate across institutional boundaries.

Our formalisation uses situation calculus to logically describe each stage (situation) of the workflow. For example, attributes (and similarly obligations) are attached to a list $H$ of history they have been through (for disambiguation) and a situation $S$ that they apply to, in the fluent $attribute(N, T, V, H, S)$. The original data rules are fluents that hold in the initial situation $s0$.

For instance, a \emph{data rule} may dictate that field \#3 in the data is \emph{private} and any use of it should be reported to the data provider. This would involve (a) an attribute which we name $pf$ (private field), a type $column$ describing the fact that it refers to a specific field, and a value $3$ to indicate the 3rd field. and (b) an obligation requiring a $report$ action be taken when \emph{any} action is performed ($action = *$) on the bound data:
\begin{align*}
&attribute(pf, column, 3, [input1,  pf\_1], s0). \\
&obligation(report, [[input1,  pf\_1]], action = *, input1, s0).
\end{align*}
The \emph{flow rules} describe how data rules flow through the process, by propagating and manipulating data rules in order to reflect how modifications performed in the data affect the policies. For example, if a process changes the column order in one of its outputs, then the column index in the data rule is changed for that output for downstream processes. This modification is encoded as a flow rule as follows:
\begin{align*}
edit(input1, output2, *, column, 3, column, 2)
\end{align*}
This states that the process changes \emph{column} \emph{3} to \emph{column} \emph{2} for data coming in from port $input1$ and results to port $output2$, matching attributes with \emph{any} ($*$) name. The mechanism to explicitly refer to input and output ports for each process is what allows us to formulate flow rules in a MIMO setting.

The $Do$ function applies a series of flow rules to a situation. As an example, let us consider the following situation:
\begin{align*}
 s1 = Do( &pr(input1, [output1, output2]): \\
  &edit(input1,output2,*,column,3,column,2): \\
  &end([output1,output2]),s0))
\end{align*}
In this, $s1$ is the result of a \emph{propagation} rule ($pr(P_{in}, Ps_{out})$ propagates all data rules from port $P_{in}$ to every port in $Ps_{out}$), the $edit$ rule discussed above, and the end of the process with 2 outputs (one untouched and one edited). Querying is then analogous to the projection task, i.e.~querying fluents that hold at the targeted final situation. For example, the query $attribute(N, T, V, H, s1)$ yields the result:
\begin{align*}
attribute(&pf, column, 3, [output1, input1, pf\_1], s1) \\
attribute(&pf, column, 2, [output2, input1, pf\_1], s1)
\end{align*}
This example is simple, but real-life rules and workflows are lengthier, so people lose track of them for MIMO multi-staged data-flow graphs.
Because situation calculus does not support parallel actions, topological sort is used to linearize the action sequence of graph-wide reasoning.


Our implementation performs retrospective analysis, using provenance as the \textit{lingua franca} of data-flow graphs. It supports two distinct provenance schemes: CWLProv\footnote{\url{https://w3id.org/cwl/prov/}} and S-Prov\footnote{\url{https://github.com/aspinuso/s-provenance}}: CWLProv is developed by the CWL community, which is a file-oriented workflow system; S-Prov is a scheme for dispel4py, a fine-grained data-streaming workflow system. In order to support this we use an intermediate abstract representation of the data-flow graphs, and extract and convert the original provenance to that using SPARQL queries.

Our evaluation is based on two real-life scientific workflows: cyclone tracking and Moment Tensor in 3D (MT3D). We use the provenance generated by the execution of the corresponding scientific workflows, where some provide CWLProv and others S-Prov. The actual data-governance rules involved in the executions are encoded using our formal language to test our language's coverage; our framework's correctness is tested by performing reasoning over data-flow graphs extracted from provenance to obtain the activated obligations and the derived data-governance rules associated with each run's data products. 
We also encode a series of real-life data-governance rules, whether used in the experiments or not, to demonstrate the generality of the model.

\section{Future Direction}

Further work will be focused on expanding our language beyond obligations (for instance to include \emph{prohibitions} checked \emph{before} using the data), establishing a more formal link with other work (e.g.~decentralized Information Flow Control) in order to connect with a wealth of existing work that uses it, and performing static optimization of flow rules to reduce the reasoning complexity and improve efficiency.

\bibliographystyle{named}
\bibliography{references}

\end{document}